\documentclass[11pt]{article}

\usepackage[a4paper,margin=1in]{geometry}
\usepackage{graphicx}
\usepackage{amsmath,amssymb}
\usepackage{booktabs}
\usepackage{placeins}
\usepackage{url}
\usepackage{adjustbox}
\usepackage[hidelinks]{hyperref}

\begin{document}
\emergencystretch=2em
\sloppy

\title{Spectral structural distortion reveals redundant neurons in neural networks}

\author{
Yongyu Wang\\
\small ORCID: \href{https://orcid.org/0009-0006-0705-752X}{0009-0006-0705-752X}\\
\small Email: \href{mailto:yongyuw@mtu.edu}{yongyuw@mtu.edu}
}

\date{}

\maketitle

\begin{abstract}
Overparameterized neural networks often contain many removable neurons, yet what makes a neuron redundant remains poorly understood. Existing pruning criteria commonly rely on local quantities such as weight magnitude, activation strength, or gradient sensitivity, but these measures provide limited insight into the structural role of a neuron in the transformation performed by a layer. Here we show that neuronal redundancy can be characterized by weak participation in the spectral structural distortion induced by layer-wise representation transformations. For each hidden layer of a trained network, we record pre-activation and post-activation hidden states, model neurons as graph nodes, and construct input-side and output-side graphs that describe neuron-level relational structure before and after the layer transformation. We then define a spectral structural importance score that measures the contribution of each neuron to the dominant graph-spectral distortion between these two relational structures. Low-participation neurons are treated as structurally redundant and removed through an iterative pruning process in which scores are recomputed after each structural change. No parameter updates are performed during intermediate pruning rounds; after the target parameter reduction is reached, a single recovery fine-tuning stage is applied to the compact model. Direct ablation analysis and experiments across conventional neural networks, encoder-only Transformers, and decoder-only language models show that this graph-spectral criterion identifies removable neurons and Transformer units while preserving task performance after compression. These results suggest that neural redundancy is not merely a consequence of small weights or weak activations, but can be understood through weak participation in the spectral distortion of layer-wise relational structure.
\end{abstract}


\section{Introduction}

Modern neural networks are often highly overparameterized: they contain far more parameters, neurons, and structural units than appear to be strictly necessary for solving a given task. This overparameterization has been central to the success of deep learning, enabling neural networks to learn complex nonlinear transformations, hierarchical representations, and flexible input-output mappings across domains ranging from computer vision to natural language processing \cite{Schmidhuber2014DeepLI,lecun2015deep,Vaswani2017AttentionIA}. At the same time, the existence of large numbers of seemingly removable parameters raises a fundamental question: what makes a neuron redundant in a trained neural network?

This question is important not only for model compression, but also for understanding how neural networks internally organize and transform information. Existing evidence from pruning, sparsity, and efficient model design shows that many weights, neurons, channels, or Transformer units can be removed with limited loss of performance \cite{molchanov2019importance,hoefler2021sparsity,blalock2020state,Muralidharan2024CompactLM}. However, the criteria used to identify removable components are often based on local or approximate quantities, such as weight magnitude, activation strength, gradient sensitivity, or task-specific loss changes. Although these criteria can be effective in practice, they do not directly explain the structural role of a neuron in the internal transformation performed by a layer. A neuron may have small weights but still participate in an important relational transformation, or it may show non-negligible activation while contributing little to the dominant structural change induced by the layer. Thus, neuronal redundancy cannot be fully understood from local scalar quantities alone.

Here we study neuronal redundancy from a graph-spectral perspective. Instead of treating neurons as isolated units, we regard the neurons in each hidden layer as nodes in a relational structure. For a trained network, we record the hidden states of each layer over inference samples and use these neuron-level observation patterns to infer two graphs: an input-side graph constructed from pre-activation states and an output-side graph constructed from post-activation states. These two graphs describe how neurons are related before and after the layer transformation. From this viewpoint, a layer does not merely transform individual neuron activations; it also deforms the relational structure among neurons. The key idea of this work is that a neuron is structurally important if it strongly participates in the dominant spectral distortion from the input-side graph to the output-side graph, and structurally redundant if it participates only weakly in this distortion.

Based on this idea, we define a neuron-level spectral structural importance score. For each hidden layer, we analyze the dominant graph-spectral directions associated with the distortion between the input-side and output-side neuron graphs. The contribution of each neuron is then estimated by aggregating the distortion scores of its local graph relationships. Neurons with low scores are interpreted as weak participants in the layer-wise structural transformation and are selected for removal. Importantly, pruning is performed iteratively: after a small portion of low-importance neurons is removed, the graph structure and neuron importance scores are recomputed under the updated network architecture. This allows the method to track how redundancy changes as the model structure becomes progressively smaller. No parameter updates are performed during the intermediate pruning rounds; after the target parameter reduction is reached, a single recovery fine-tuning stage is applied to adapt the compact model.

This framework provides a structural interpretation of neural network redundancy. It suggests that removable neurons are not simply those with small weights, weak activations, or low local sensitivity, but those that contribute little to the dominant spectral deformation of layer-wise relational structure. In this sense, pruning becomes a way to test whether weak participation in spectral structural distortion is a reliable indicator of neuronal redundancy. We evaluate this principle across conventional neural networks, encoder-only Transformer models, and decoder-only large language models. The results show that the proposed graph-spectral criterion can identify removable neurons, channels, feed-forward intermediate units, and attention head subspaces across diverse architectures while preserving task performance after compression. These findings support the view that structural redundancy in neural networks can be characterized through the spectral geometry of layer-wise representation transformations.

\section{Methods}\label{sec:methods}

\subsection{Graph-spectral view of neuronal redundancy}

The goal of the proposed method is to identify neurons or structural units that contribute weakly to the layer-wise transformation of internal representation structure. We use the term neuron in a broad sense to denote a prunable computational unit, including a neuron in a fully connected layer, a channel in a convolutional layer, a feed-forward intermediate unit in a Transformer block, or an attention head subspace when the same scoring principle is applied to attention modules.

\begin{figure}[t!]
\centering
\includegraphics[width=0.90\textwidth]{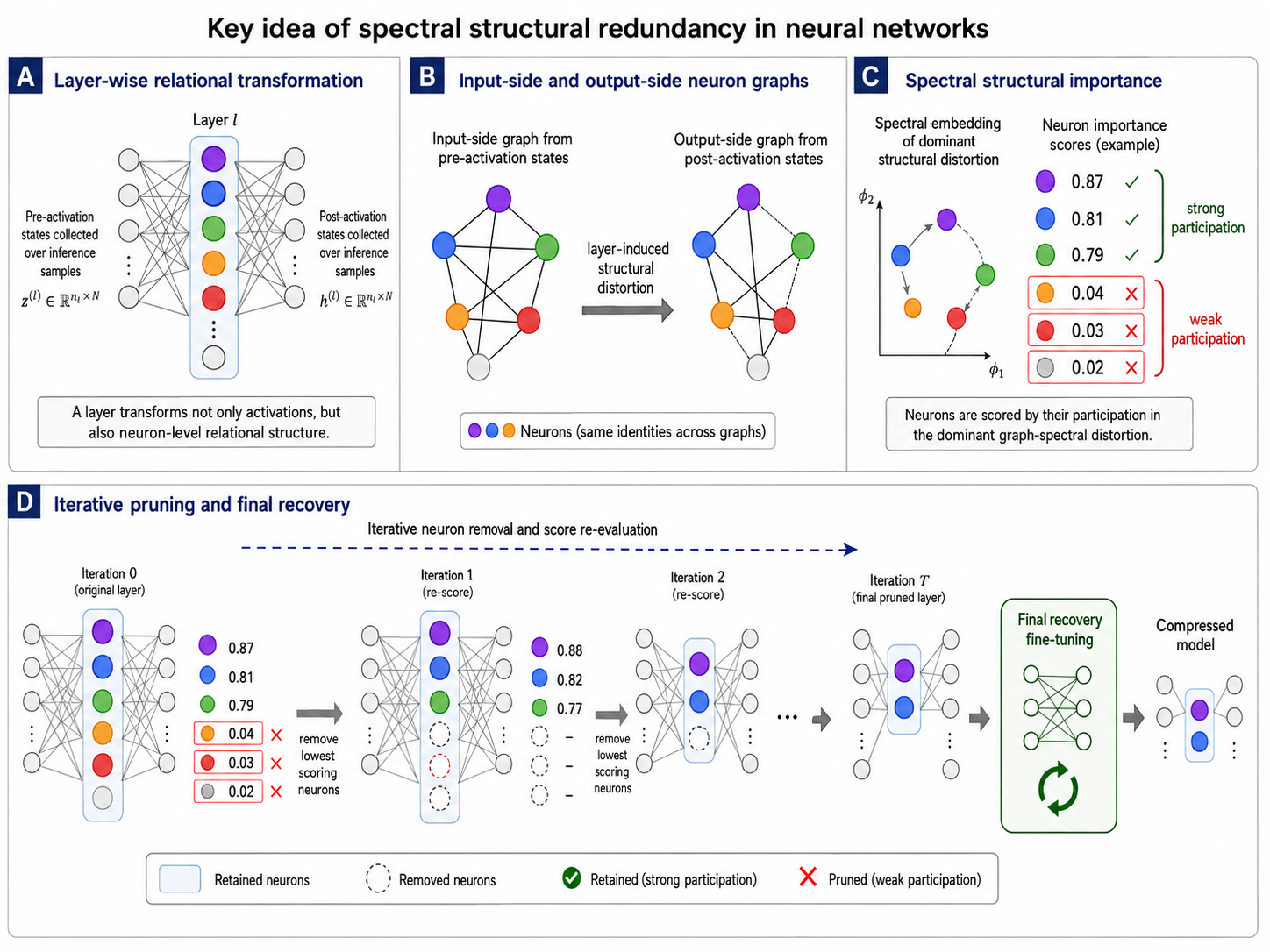}
\caption{Key idea of spectral structural redundancy in neural networks. Hidden states are used to construct input-side and output-side neuron graphs. The layer transformation is interpreted as a deformation of neuron-level relational structure, and each neuron is scored according to its participation in the dominant graph-spectral distortion. Low-participation neurons are removed iteratively, followed by a single recovery fine-tuning stage.}
\label{fig:Fig1}
\end{figure}

The central idea is illustrated in Figure~\ref{fig:Fig1}. For each hidden layer of a trained network, we collect pre-activation and post-activation hidden states over a set of inference samples. These hidden-state observations are used to infer two neuron-level graphs: an input-side graph describing the relational structure among neurons before the layer transformation, and an output-side graph describing the relational structure after the layer transformation. The layer is then interpreted as inducing a deformation from the input-side relational structure to the output-side relational structure. We quantify this deformation using a graph-spectral distortion analysis and assign each neuron a spectral structural importance score according to its participation in the dominant distortion directions. Neurons with weak participation are treated as structurally redundant and are removed iteratively, with scores recomputed after each structural change. After the target parameter reduction is reached, a single recovery fine-tuning stage is applied to the compact model.

\subsection{Hidden-state collection}

Let \(f_{\theta}\) denote a trained neural network and let \(\mathcal{C}=\{x_k\}_{k=1}^{n}\) be a set of inference samples used for structural estimation. These samples are used only to collect hidden-state statistics and to estimate neuron-level relational structure; held-out test data are not used for pruning-score estimation.

For the \(l\)-th hidden layer, let \(m_l\) denote the number of neurons or prunable structural units. During a forward pass on sample \(x_k\), we record for each unit \(i\) its pre-activation input and post-activation output, denoted by \(z_i^{(l,k)}\) and \(h_i^{(l,k)}\), respectively. After running the network over all samples, each unit is represented by
\begin{equation}
\mathbf{z}_i^{(l)}=
\big(z_i^{(l,1)},\ldots,z_i^{(l,n)}\big)^\top,
\end{equation}
and
\begin{equation}
\mathbf{h}_i^{(l)}=
\big(h_i^{(l,1)},\ldots,h_i^{(l,n)}\big)^\top .
\end{equation}
The vector \(\mathbf{z}_i^{(l)}\) describes how the input received by unit \(i\) varies across inference samples, whereas \(\mathbf{h}_i^{(l)}\) describes how the output produced by the same unit varies after the layer transformation.

Before graph construction, each observation vector is standardized across inference samples:
\begin{equation}
\widetilde{\mathbf{z}}_i^{(l)}=
\frac{\mathbf{z}_i^{(l)}-\mu_i^z\mathbf{1}}
{\sigma_i^z+\epsilon},\qquad
\widetilde{\mathbf{h}}_i^{(l)}=
\frac{\mathbf{h}_i^{(l)}-\mu_i^h\mathbf{1}}
{\sigma_i^h+\epsilon}.
\end{equation}
Here \(\mu_i^z\) and \(\sigma_i^z\) are the empirical mean and standard deviation of \(\mathbf{z}_i^{(l)}\), and \(\mu_i^h\) and \(\sigma_i^h\) are defined analogously for \(\mathbf{h}_i^{(l)}\). The constant \(\epsilon\) is used for numerical stability. This standardization reduces scale differences among units so that graph construction mainly reflects similarity in response patterns rather than raw activation magnitude.

\subsection{Construction of input-side and output-side neuron graphs}

For each hidden layer, we construct two graphs with the same vertex set:
\begin{equation}
G_{\rm in}^{(l)}=(\mathcal{V}^{(l)},\mathcal{E}_{\rm in}^{(l)},\mathbf{W}_{\rm in}^{(l)}),
\end{equation}
and
\begin{equation}
G_{\rm out}^{(l)}=(\mathcal{V}^{(l)},\mathcal{E}_{\rm out}^{(l)},\mathbf{W}_{\rm out}^{(l)}).
\end{equation}
Here \(\mathcal{V}^{(l)}=\{1,\ldots,m_l\}\) denotes the set of neurons or structural units in layer \(l\). The input-side graph is constructed from standardized pre-activation vectors, and the output-side graph is constructed from standardized post-activation vectors.

For the input-side graph, pairwise distances are computed as
\begin{equation}
d_{\rm in}^{(l)}(i,j)=
\big\|\widetilde{\mathbf{z}}_i^{(l)}-
\widetilde{\mathbf{z}}_j^{(l)}\big\|_2 .
\end{equation}
A symmetric \(k\)-nearest-neighbor graph is formed by connecting two units if either unit is among the \(k\) nearest neighbors of the other. The edge weight is defined using a Gaussian kernel:
\begin{equation}
W_{\rm in}^{(l)}(i,j)=
\exp\!\left[-\frac{(d_{\rm in}^{(l)}(i,j))^2}{2(\tau_{\rm in}^{(l)})^2}\right]
\end{equation}
for connected pairs \((i,j)\), and is set to zero otherwise. The output-side graph is constructed analogously from the post-activation vectors, yielding \(\mathcal{E}_{\rm out}^{(l)}\) and \(\mathbf{W}_{\rm out}^{(l)}\).

The input-side graph captures the relational structure among units before the layer transformation, while the output-side graph captures the relational structure after the transformation. For each graph, we compute the unnormalized graph Laplacian:
\begin{equation}
\mathbf{L}_{\rm in}^{(l)}=
\mathbf{D}_{\rm in}^{(l)}-\mathbf{W}_{\rm in}^{(l)},
\end{equation}
and
\begin{equation}
\mathbf{L}_{\rm out}^{(l)}=
\mathbf{D}_{\rm out}^{(l)}-\mathbf{W}_{\rm out}^{(l)},
\end{equation}
where \(\mathbf{D}_{\rm in}^{(l)}\) and \(\mathbf{D}_{\rm out}^{(l)}\) are the corresponding degree matrices. These Laplacians provide the basis for measuring how graph signals vary over the input-side and output-side relational structures.

\subsection{Spectral structural distortion}

To quantify the structural deformation induced by layer \(l\), we consider a graph signal \(\mathbf{x}\in\mathbb{R}^{m_l}\) defined over the units of that layer. Its variation on the input-side and output-side graphs is measured by the quadratic forms \(\mathbf{x}^\top\mathbf{L}_{\rm in}^{(l)}\mathbf{x}\) and \(\mathbf{x}^\top\mathbf{L}_{\rm out}^{(l)}\mathbf{x}\). We define the relative graph variation as
\begin{equation}
R^{(l)}(\mathbf{x})=
\frac{\mathbf{x}^\top\mathbf{L}_{\rm in}^{(l)}\mathbf{x}}
{\mathbf{x}^\top\mathbf{L}_{\rm out}^{(l)}\mathbf{x}} .
\end{equation}
This quotient identifies directions whose graph variation differs strongly between the input-side and output-side structures. According to the generalized Courant--Fischer theorem~\cite{Golub1983MatrixC}, the dominant values are obtained from
\begin{equation}
\mathbf{L}_{\rm in}^{(l)}\mathbf{v}=
\lambda\mathbf{L}_{\rm out}^{(l)}\mathbf{v} .
\end{equation}
Equivalently, the dominant generalized directions can be obtained from the leading eigenvectors of \((\mathbf{L}_{\rm out}^{(l)})^+\mathbf{L}_{\rm in}^{(l)}\), where \((\cdot)^+\) denotes the Moore--Penrose pseudoinverse.

We compute the top \(s\) generalized eigenvalues \(\lambda_1^{(l)},\ldots,\lambda_s^{(l)}\) and their corresponding eigenvectors \(\mathbf{v}_1^{(l)},\ldots,\mathbf{v}_s^{(l)}\). Following the idea of dominant generalized-eigenvector embeddings~\cite{cheng2021spade}, we define the weighted spectral embedding compactly as
\begin{equation}
\mathbf{V}_s^{(l)}=
\big[\sqrt{\lambda_r^{(l)}}\,\mathbf{v}_r^{(l)}\big]_{r=1}^{s}.
\end{equation}
The \(i\)-th row \(\mathbf{V}_s^{(l)}(i,:)\) represents unit \(i\) in the dominant spectral distortion coordinates of layer \(l\).

\subsection{Neuron-level spectral structural importance}

The spectral embedding allows us to measure how strongly each local neuron relationship participates in the dominant structural distortion. For an edge \((i,j)\in\mathcal{E}_{\rm in}^{(l)}\), we define the edge-level distortion score as
\begin{equation}
S_{\rm edge}^{(l)}(i,j)=
\big\|\mathbf{V}_s^{(l)}(i,:)-
\mathbf{V}_s^{(l)}(j,:)\big\|_2^2 .
\end{equation}
A larger edge score indicates that the local relationship between units \(i\) and \(j\) is strongly aligned with the dominant graph-spectral distortion induced by the layer transformation.

To obtain a neuron-level score, we aggregate edge-level scores over the input-side neighborhood. Let \(\mathcal{N}_{\rm in}^{(l)}(i)\) denote the neighbor set of unit \(i\). The spectral structural importance score is
\begin{equation}
S_{\rm node}^{(l)}(i)=
\frac{1}{|\mathcal{N}_{\rm in}^{(l)}(i)|}
\sum_{j\in\mathcal{N}_{\rm in}^{(l)}(i)}
S_{\rm edge}^{(l)}(i,j).
\end{equation}
This score measures the extent to which a unit participates in the dominant spectral structural distortion from the input-side graph to the output-side graph. Units with high scores are interpreted as structurally important because their local relationships are strongly involved in the layer-induced deformation. Units with low scores are interpreted as structurally redundant because their local relationships participate weakly in the dominant deformation and are therefore safer candidates for pruning.

\subsection{Iterative pruning with score re-evaluation}

The importance score is computed under the current network structure. Because removing units changes the remaining relational structure, we do not prune all low-scoring units in a single step. Instead, for a target effective parameter reduction ratio \(\rho\), the pruning process is divided into \(T\) iterations.

Let \(M_0\) be the original trained model and \(M_t\) be the model after the \(t\)-th pruning iteration. The effective parameter reduction after iteration \(t\) is
\begin{equation}
\rho_t=1-\frac{P(M_t)}{P(M_0)},
\end{equation}
where \(P(M)\) denotes the number of trainable parameters in model \(M\). At iteration \(t\), we collect hidden-state observations using the current model \(M_{t-1}\), reconstruct the input-side and output-side graphs, recompute the spectral structural importance scores, and remove a small set of low-scoring units subject to the target pruning budget and architectural constraints. The cumulative pruning budget is chosen so that \(\rho_t\) gradually approaches the target ratio \(\rho\).

This iterative procedure has two purposes. First, it avoids making a large structural change based on scores computed only from the original network. Second, it allows the importance of remaining units to be re-evaluated after each pruning step, so that redundancy is estimated relative to the current compact architecture rather than a fixed initial architecture.

During the iterative pruning stage, the remaining parameters are inherited directly from the previous model and are not updated by gradient descent. Thus, the pruning stage measures structural redundancy without allowing intermediate fine-tuning to compensate for each removal step. Once the target parameter reduction is reached, the final compact model inherits the surviving parameters from the original trained model and is fine-tuned once on the downstream task. This final recovery stage adapts the remaining parameters to the compact architecture but is not used to select pruning candidates.

\subsection{Application to different neural architectures}

The same graph-spectral scoring principle can be applied to different neural architectures by defining the prunable unit appropriately. In fully connected layers, each hidden neuron is treated as a graph node. In convolutional networks, each channel or filter-level unit can be treated as a node by collecting its response pattern over inference samples. In Transformer feed-forward modules, each intermediate feed-forward unit is treated as a node. For attention modules, a head or head subspace can be treated as a structural unit when the model architecture permits structured head-level pruning.

Architectural constraints are enforced during pruning. For example, embedding layers, output prediction heads, normalization layers, and other non-prunable components can be kept intact. In decoder-only language models with grouped-query attention, key and value projections may be preserved when required to maintain architectural consistency, while feed-forward intermediate units and query/output head subspaces can be pruned. These architecture-specific choices do not change the scoring principle: in each case, the method estimates how weakly a structural unit participates in the spectral distortion of layer-wise relational structure and removes low-participation units under the allowed pruning constraints.

\section{Results}\label{sec:results}

We evaluated whether weak participation in layer-wise spectral structural distortion can serve as a general indicator of neuronal redundancy. The experiments were designed to test three questions. First, does the proposed spectral structural importance score directly relate to the functional damage caused by removing a unit? Second, can the proposed graph-spectral criterion identify removable units in conventional neural networks with different architectures and tasks? Third, can the same principle be extended to encoder-only and decoder-only Transformer language models, where the prunable units include feed-forward intermediate neurons and attention head subspaces?

Unless otherwise stated, pruning was performed without intermediate parameter updates. After the target compact architecture was obtained, a single recovery fine-tuning stage was applied. The reported performance was measured on held-out test sets that were not used for pruning-score estimation or recovery fine-tuning. We report effective parameter reduction as the actual reduction in total trainable parameters:
\[
\rho_{\mathrm{eff}}
=
1-
\frac{
P(M_{\mathrm{pruned}})
}
{
P(M_{\mathrm{original}})
},
\]
where \(P(M)\) denotes the number of trainable parameters in model \(M\). This quantity differs from the pruning-unit reduction ratio, which measures the percentage of removable structural units, such as FFN intermediate neurons or attention heads, selected for deletion.

\subsection{Spectral structural importance predicts immediate pruning damage}

A central assumption of the proposed framework is that neurons with weak participation in layer-wise spectral structural distortion are safer to remove. To directly test this assumption, we performed an immediate ablation analysis on a fully trained LeNet-5 model on MNIST. Unlike the pruning experiments reported below, this analysis did not include recovery fine-tuning. Instead, groups of neurons or channels were temporarily removed from the trained model, and the immediate validation accuracy drop was measured. This design directly evaluates whether the spectral structural importance score is associated with the functional vulnerability of removable units.

For each analyzed layer, spectral structural importance scores were computed from hidden-state observations on a calibration set. Units in the main prunable hidden layers were then divided into low-score, middle-score, and high-score groups according to their within-layer spectral structural importance percentiles. The analysis focused on the main prunable hidden layers, where neuron- or channel-level relational graphs contain sufficient units for stable spectral estimation. For each score group, small groups of units were temporarily ablated without any recovery fine-tuning, and the resulting validation accuracy drop was recorded.

\begin{figure}[t!]
\centering
\includegraphics[width=0.98\linewidth]{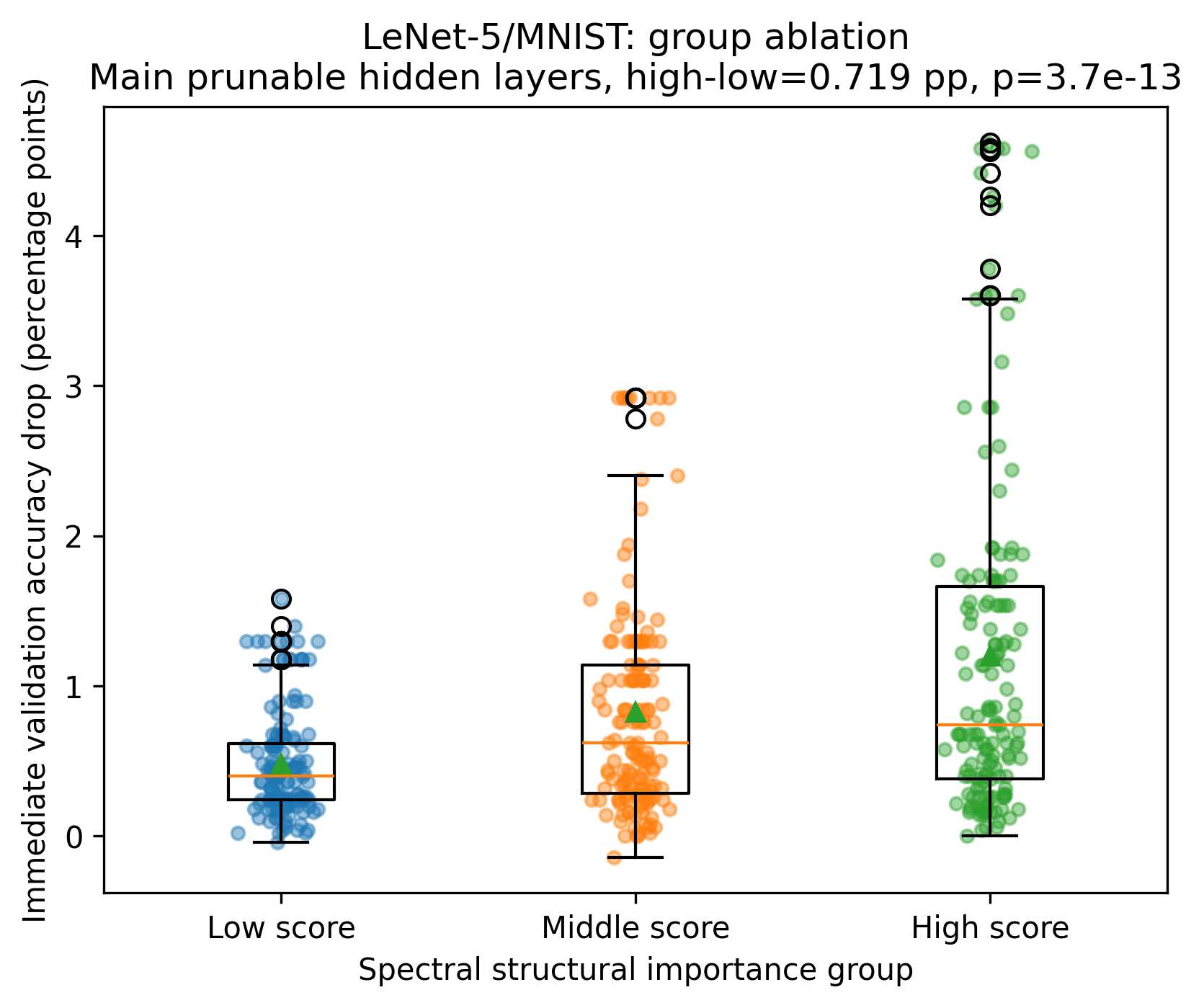}
\caption{Spectral structural importance predicts immediate group-level pruning damage. Small groups of low-, middle-, and high-score units were temporarily ablated in the main prunable hidden layers of LeNet-5 on MNIST without recovery fine-tuning. High-score unit groups caused larger immediate validation accuracy drops than low-score groups.}
\label{fig:score_damage_lenet5}
\end{figure}

As shown in Figure~\ref{fig:score_damage_lenet5}, high-score unit groups caused substantially larger immediate validation accuracy drops than low-score groups. The mean difference between high-score and low-score groups was \(0.719\) percentage points, and the paired comparison was highly significant (\(p=3.7\times10^{-13}\)). This result indicates that units with stronger participation in spectral structural distortion are more functionally vulnerable to removal, whereas low-score units are empirically safer pruning candidates.

These results provide a direct functional test of the proposed score. They show that spectral structural importance is not merely a descriptive graph quantity, but is associated with the immediate damage caused by removing groups of units from a trained network. In particular, weak participation in the dominant spectral deformation of layer-wise relational structure serves as an empirical indicator of structural redundancy.

\subsection{Spectral structural redundancy in conventional neural networks}

We next examined whether the proposed criterion identifies redundant neurons or channels in conventional neural networks. The evaluation included image classification, reconstruction, sensor-based classification, and medical image segmentation tasks. Specifically, we tested LeNet-5 on MNIST and Fashion-MNIST, LeNet-300-100 on MNIST and Fashion-MNIST, an autoencoder on MNIST reconstruction, an MLP on UCI HAR human activity classification, AlexNet on CIFAR-10, and U-Net on Kvasir-SEG polyp segmentation.

These models cover several forms of neural computation. LeNet-5 and AlexNet represent convolutional image classifiers, LeNet-300-100 and the MLP represent fully connected networks, the autoencoder tests whether the method applies to reconstruction rather than classification, and U-Net tests whether the criterion remains useful in a dense prediction setting. Across these settings, the target effective parameter reduction was approximately 50\%.

For the smaller models, including LeNet-5, LeNet-300-100, the autoencoder, and the MLP, the original models and the pruned models were trained using the same one-epoch budget. For AlexNet on CIFAR-10 and U-Net on Kvasir-SEG, larger training budgets were used because these tasks require more training before pruning becomes meaningful. AlexNet was trained for 5 epochs before pruning and recovered for 5 epochs after pruning. U-Net was trained for 30 epochs before pruning and recovered for 10 epochs after pruning.

As shown in Table~\ref{tab:nn_pruning_results}, the proposed method reduced model size by approximately 50\% across all conventional networks while preserving task performance. In several classification settings, the compact models slightly improved test accuracy after recovery fine-tuning. The MNIST autoencoder maintained comparable reconstruction error, and the pruned U-Net preserved Dice and IoU scores after compression. These results suggest that low participation in spectral structural distortion identifies removable units across convolutional, fully connected, reconstruction-based, and segmentation architectures.

\begin{table}[t]
\centering
\caption{Performance and compression results on conventional neural networks. Effective Param. Red. denotes the actual reduction in total trainable parameters. Acc. denotes accuracy; Cls., Rec., and Seg. denote classification, reconstruction, and segmentation.}
\label{tab:nn_pruning_results}
\footnotesize
\setlength{\tabcolsep}{5pt}
\renewcommand{\arraystretch}{1.16}
\begin{adjustbox}{width=\textwidth}
\begin{tabular}{l l l c c c c}
\toprule
Model & Dataset & Task & Metric
& Params. & Effective Param. Red. & Test Set \\
\midrule
LeNet-5 & MNIST & Cls. & Acc.
& 44,426 $\rightarrow$ 20,735
& 53.33\%
& 93.34\% $\rightarrow$ 94.09\% \\

LeNet-5 & Fashion-MNIST & Cls. & Acc.
& 44,426 $\rightarrow$ 20,158
& 54.63\%
& 74.51\% $\rightarrow$ 75.13\% \\

LeNet-300-100 & MNIST & Cls. & Acc.
& 266,610 $\rightarrow$ 131,823
& 50.56\%
& 94.40\% $\rightarrow$ 95.24\% \\

LeNet-300-100 & Fashion-MNIST & Cls. & Acc.
& 266,610 $\rightarrow$ 123,822
& 53.56\%
& 83.92\% $\rightarrow$ 84.70\% \\

Autoencoder & MNIST & Rec. & MSE
& 435,536 $\rightarrow$ 216,504
& 50.29\%
& 0.035295 $\rightarrow$ 0.034572 \\

MLP & UCI HAR & Cls. & Acc.
& 80,582 $\rightarrow$ 40,102
& 50.23\%
& 83.41\% $\rightarrow$ 85.17\% \\

AlexNet & CIFAR-10 & Cls. & Acc.
& 947,146 $\rightarrow$ 392,087
& 58.60\%
& 68.60\% $\rightarrow$ 71.69\% \\

U-Net & Kvasir-SEG & Seg. & Dice/IoU
& 118,273 $\rightarrow$ 55,936
& 52.71\%
& 0.62/0.44 $\rightarrow$ 0.63/0.46 \\
\bottomrule
\end{tabular}
\end{adjustbox}
\end{table}
\FloatBarrier

\subsection{Extension to encoder-only Transformer language models}

We next tested whether the graph-spectral redundancy criterion can be applied to Transformer language models. We first evaluated BERT-base on six GLUE tasks: SST-2, QNLI, MRPC, QQP, MNLI, and CoLA. These tasks cover sentiment classification, natural language inference, paraphrase detection, duplicate-question detection, and linguistic acceptability judgment. Accuracy was used for SST-2, QNLI, and MNLI; F1 score was used for MRPC and QQP; and Matthews correlation coefficient was used for CoLA.

In each BERT Transformer block, the feed-forward network contains intermediate neurons and the self-attention module contains multiple attention heads. We therefore treated FFN intermediate neurons and attention heads as prunable structural units. The pruning-unit reduction ratio was set to approximately 40\%, and pruning was divided into 10 iterations. At each iteration, spectral structural importance scores were recomputed under the current compact architecture. The original BERT-base model was fine-tuned for one epoch before pruning, and the pruned model was recovered using one epoch of LoRA-based fine-tuning.

Table~\ref{tab:bert_glue_pruning_results} shows that removing approximately 40\% of prunable Transformer units led to an effective total parameter reduction of approximately 25--27\%, depending on the task-specific classification head and pruning allocation. After LoRA recovery, the compact BERT models preserved competitive performance on several tasks. The pruned model slightly improved performance on QNLI and MRPC, showed moderate degradation on SST-2, QQP, and MNLI, and showed a larger degradation on CoLA. This pattern suggests that the recoverability of structurally pruned Transformer units depends on the downstream task. In particular, linguistic acceptability judgment appears more sensitive to structural removal under the current recovery setting.

\begin{table}[t]
\centering
\caption{Performance and compression results on BERT-base for GLUE tasks. Unit Red. denotes the reduction ratio of prunable FFN intermediate neurons and attention heads. Effective Param. Red. denotes the actual reduction in total trainable parameters.}
\label{tab:bert_glue_pruning_results}
\footnotesize
\setlength{\tabcolsep}{3.6pt}
\renewcommand{\arraystretch}{1.15}
\begin{adjustbox}{width=\textwidth}
\begin{tabular}{l l l c c c c c}
\toprule
Model & Dataset & Task & Metric
& Params. & Unit Red. & Effective Param. Red. & Test Set \\
\midrule
BERT-base & SST-2 & Sentiment Cls. & Acc.
& 109,483,778 $\rightarrow$ 79,860,474
& 40.06\% & 27.06\%
& 87.39\% $\rightarrow$ 85.89\% \\

BERT-base & QNLI & NLI & Acc.
& 109,483,778 $\rightarrow$ 82,418,874
& 40.03\% & 24.72\%
& 77.10\% $\rightarrow$ 77.90\% \\

BERT-base & MRPC & Paraphrase Detection & F1
& 109,483,778 $\rightarrow$ 81,828,474
& 40.04\% & 25.26\%
& 86.23\% $\rightarrow$ 86.96\% \\

BERT-base & QQP & Duplicate Question & F1
& 109,483,778 $\rightarrow$ 80,450,874
& 40.06\% & 26.52\%
& 73.21\% $\rightarrow$ 70.96\% \\

BERT-base & MNLI & NLI & Acc.
& 109,484,547 $\rightarrow$ 82,419,643
& 40.03\% & 24.72\%
& 63.70\% $\rightarrow$ 60.90\% \\

BERT-base & CoLA & Acceptability & MCC
& 109,483,778 $\rightarrow$ 82,222,074
& 40.03\% & 24.90\%
& 0.5590 $\rightarrow$ 0.3955 \\
\bottomrule
\end{tabular}
\end{adjustbox}
\end{table}
\FloatBarrier

\subsection{Physical compression of decoder-only language models}

Finally, we evaluated whether the proposed criterion can be extended to decoder-only language models and produce real physical compression. We tested TinyLlama-1.1B-Chat and Qwen2.5-0.5B on SST-2 under a generative classification setting, where the model predicts the sentiment label for each input sentence.

Both models use grouped-query attention. To preserve architectural consistency, we kept the token embedding, final normalization, key/value projections, and language modeling head unchanged. We physically pruned FFN intermediate neurons and query/output attention head subspaces. This setting is more restrictive than unstructured sparsity because the resulting model has fewer actual parameters and a smaller architecture.

For each model, we evaluated three pruning levels with increasing reductions in prunable FFN and Q/O attention units. The pruning process was divided into 3 iterations. The original model was first fine-tuned using LoRA for 3 epochs. After pruning, the compact model was recovered using LoRA fine-tuning for another 3 epochs on the same downstream task. The LoRA rank was set to 64 and the LoRA scaling factor was set to 128. Evaluation was performed on a held-out test set that was not used for pruning calibration or recovery fine-tuning.

As shown in Table~\ref{tab:tinyllama_qwen_sst2_pruning_results}, the method achieved substantial physical parameter reduction in both decoder-only models. For TinyLlama-1.1B-Chat, the strongest setting reduced the model from 1.10B to 517.74M parameters, corresponding to 52.93\% effective parameter reduction, while SST-2 accuracy changed from 93.00\% to 87.73\%. Notably, the performance remained similar between the medium and strongest pruning levels, despite a large increase in effective parameter reduction from 37.03\% to 52.93\%. This suggests that, for this task, many additional FFN and Q/O attention units contribute limited recoverable utility after the main redundant structure has been removed.

For Qwen2.5-0.5B, the model also maintained moderate performance under the lower pruning levels, but the strongest setting caused a larger degradation. This indicates that the compression tolerance of decoder-only models is architecture- and task-dependent. Overall, these results show that weak participation in spectral structural distortion can identify removable units not only in conventional neural networks and encoder-only Transformers, but also in decoder-only language models with physical architecture reduction.

\begin{table}[t]
\centering
\caption{Physical pruning results on TinyLlama-1.1B-Chat and Qwen2.5-0.5B for SST-2. Unit Red. denotes the reduction ratio of prunable FFN intermediate neurons and query/output attention head subspaces. Effective Param. Red. denotes actual parameter reduction after physical pruning.}
\label{tab:tinyllama_qwen_sst2_pruning_results}
\footnotesize
\setlength{\tabcolsep}{3.5pt}
\renewcommand{\arraystretch}{1.15}
\begin{adjustbox}{width=\textwidth}
\begin{tabular}{l l l c c c c c}
\toprule
Model & Dataset & Task & Metric
& Params. & Unit Red. & Effective Param. Red. & Test Set \\
\midrule
TinyLlama-1.1B-Chat & SST-2 & Sentiment Cls. & Acc.
& 1,100,048,384 $\rightarrow$ 917,211,136
& 18.73\% & 16.62\%
& 93.00\% $\rightarrow$ 90.48\% \\

TinyLlama-1.1B-Chat & SST-2 & Sentiment Cls. & Acc.
& 1,100,048,384 $\rightarrow$ 692,652,032
& 42.18\% & 37.03\%
& 93.00\% $\rightarrow$ 87.84\% \\

TinyLlama-1.1B-Chat & SST-2 & Sentiment Cls. & Acc.
& 1,100,048,384 $\rightarrow$ 517,744,640
& 60.61\% & 52.93\%
& 93.00\% $\rightarrow$ 87.73\% \\

Qwen2.5-0.5B & SST-2 & Sentiment Cls. & Acc.
& 494,032,768 $\rightarrow$ 427,000,192
& 18.74\% & 13.57\%
& 91.86\% $\rightarrow$ 89.33\% \\

Qwen2.5-0.5B & SST-2 & Sentiment Cls. & Acc.
& 494,032,768 $\rightarrow$ 342,504,832
& 42.17\% & 30.67\%
& 91.86\% $\rightarrow$ 88.76\% \\

Qwen2.5-0.5B & SST-2 & Sentiment Cls. & Acc.
& 494,032,768 $\rightarrow$ 279,129,472
& 60.60\% & 43.50\%
& 91.86\% $\rightarrow$ 82.34\% \\
\bottomrule
\end{tabular}
\end{adjustbox}
\end{table}
\FloatBarrier

\subsection{Summary of empirical findings}

Across all evaluated architectures, the proposed graph-spectral criterion consistently identified removable structural units. In conventional neural networks, approximately half of the parameters could be removed while preserving classification, reconstruction, and segmentation performance. In BERT-base, deleting approximately 40\% of prunable Transformer units produced compact models with approximately 25--27\% actual parameter reduction and competitive performance on several GLUE tasks. In decoder-only language models, the same principle produced physically smaller TinyLlama and Qwen models by removing FFN intermediate neurons and Q/O attention head subspaces.

Together, the direct ablation analysis and compression experiments support the central hypothesis of this study: neurons or structural units that weakly participate in the dominant spectral distortion of layer-wise relational structure are more likely to be redundant. The degree to which performance can be recovered after pruning depends on architecture, task sensitivity, and recovery budget, but the same graph-spectral scoring principle remains applicable across substantially different model families.

\FloatBarrier

\section{Conclusion}

This study introduces a graph-spectral view of neuronal redundancy in neural networks. Instead of defining redundancy only through local quantities such as weight magnitude, activation strength, or sensitivity, we characterize each neuron by its participation in the spectral structural distortion between the input-side and output-side relational structures of a layer. Under this view, a layer is not only a nonlinear transformation of activations, but also a deformation of neuron-level relational geometry. Neurons that weakly participate in the dominant spectral deformation are interpreted as structurally redundant and can be selected as safer pruning candidates.

Using this principle, we developed an iterative pruning framework that constructs layer-wise neuron graphs from hidden-state observations, computes spectral structural importance scores, removes low-participation units, and recomputes importance after each structural change. Experiments across conventional neural networks, encoder-only Transformers, and decoder-only language models show that the same graph-spectral criterion can identify removable neurons, channels, feed-forward intermediate units, and attention head subspaces while preserving task performance after recovery fine-tuning. These findings suggest that neural redundancy is not merely a consequence of small weights or weak activations, but can be understood through weak participation in the spectral structural distortion induced by layer-wise representation transformations. Future work will further examine the relationship between spectral structural importance and direct pruning damage, extend the framework to larger foundation models and multimodal architectures, and investigate whether spectral structural redundancy can help characterize the limits of neural network compression.

\bibliographystyle{plain}
\bibliography{pnas-sample}

\end{document}